\documentclass{article}

\PassOptionsToPackage{numbers, compress}{natbib}

 \usepackage[preprint]{nips_2018}

\usepackage[T1]{fontenc}    
\usepackage{url}            
\usepackage{booktabs}       
\usepackage{amsfonts}       
\usepackage{nicefrac}       
\usepackage{microtype}      
\usepackage{tabularx}
\usepackage{algorithm}
\usepackage{algorithmic}
\usepackage{epsfig}
\usepackage{amsmath}
\usepackage{multirow}
\usepackage{subfigure}

\title{Graph Diffusion-Embedding Networks}

\author{
  Bo Jiang, Doudou Lin, Jin Tang\\
  School of Computer Science and Technology \\
  Anhui University\\
  Hefei, China \\
  \texttt{jiangbo@ahu.edu.cn} \\
}

\begin{document}

\maketitle

\begin{abstract}

We present a novel graph diffusion-embedding networks (GDEN) for graph-structured data.
GDEN is motivated by our closed-form formulation on \emph{regularized feature diffusion} on graph.
GDEN integrates both \emph{regularized feature diffusion} and \emph{low-dimensional embedding} simultaneously in a unified network model.
Moreover, based on GDEN, we can naturally deal with the structured data with multiple graph structures.
Experiments on semi-supervised learning tasks on several benchmark datasets demonstrate the  better performance of the proposed GDEN when
comparing with the traditional GCN models.

\end{abstract}

\section{Introduction}

\subsection{Graph based feature diffusion}

Given a graph $G(V, E)$ with $V$ denoting the $n$ nodes and $E$ representing the edges.
Let $A \in \mathbb{R}^{n\times n}$ be the corresponding adjacency matrix and $X=(X_1 \cdots X_n)\in \mathbb{R}^{n\times d}$ be the feature set of nodes,
where $X_i$ denotes the attribute vector for  node $v_i\in V$.
The aim of our graph based feature diffusion is to
learn a feature representation $Z=\mathcal{H}_d(A,X)=(Z_1\cdots Z_n)$ for each node by incorporating the contextual information of the other node representations.
In the following, we provide three kinds of graph based feature diffusion model $\mathcal{H}_d(A,X)$, as summarized in Table 1.
Similar diffusion models have been commonly used in ranking and label propagation process
~\cite{zhou2004learning,zhou2004ranking,donoser2013diffusion,lu2014learning}. Differently, in this paper, we propose to explore
them for \emph{feature diffusion} problem whose aim is to learn a contextual feature representation for each graph node.

\textbf{(1) Graph Laplacian diffusion}

Motivated by manifold ranking~\cite{zhou2004learning}, we propose to compute the optimal diffused representation $Z$ by solving the
following optimization problem:
\begin{equation}\label{EQ:Diffusion}
\min_{Z} \ \ \frac{1}{2} \sum^n_{i,j=1} A_{ij} \|Z_i - Z_j\|^2_F + \alpha \sum^n_{i=1} \|Z_i - X_i\|^2_F
\end{equation}
where $Z_i$ denotes the diffused feature of node $v_i$. 
The first term conducts feature diffusion/propagation on graph while the second term encourages to preserve the original feature information $X$ in diffusion process. 
It is known that, the optimal closed-form solution for this problem is given by
\begin{equation}\label{EQ:Diffusion}
Z^* = \alpha(I + \alpha (D - A))^{-1}X
\end{equation}
where $D = \text{diag}\{d_{11},d_{22},\cdots d_{nn}\}$ and $d_{ii} = \sum_{j}A_{ij}$, and $D - A$ is the Laplacian  of  graph.

\textbf{(2) Graph normalized Laplacian diffusion}

One can also compute the optimal diffused representation $Z$ by solving the following normalized optimization problem ~\cite{zhou2004ranking,lu2014learning}:
\begin{equation}\label{EQ:Diffusion}
\min_{Z} \ \ \frac{1}{2} \sum^n_{i,j=1} A_{ij} \|\frac{Z_i}{\sqrt{d_{ii}}} - \frac{Z_j}{\sqrt{d_{jj}}}\|^2_F + \lambda \sum^n_{i=1} \|Z_i - X_i\|^2_F
\end{equation}
The optimal closed-form solution for this problem is given by
\begin{equation}\label{EQ:Diffusion}
Z^* = \alpha(I - \alpha D^{-1/2}AD^{-1/2})^{-1}X
\end{equation}
where $D^{-1/2}AD^{-1/2}$ is the normalized Laplacian of the graph and $\alpha = 1/(1+\lambda)$.

\textbf{(3) Graph random walk diffusion}

Another method to formulate the feature diffusion is based on random walk with restart (RWR) model ~\cite{zhou2004ranking,lu2014learning} and obtain the equilibrium
  representation on graph.
In order to do so,
we first define a transition probability matrix as
\begin{equation}\label{EQ:Diffusion}
P = AD^{-1}
\end{equation}
Then, the RWR is conducted on graph and converges to an
equilibrium distribution $Z^*$.
Formally, it conducts update as
\begin{equation}\label{EQ:Diffusion}
{Z}^{t+1} = \alpha P Z^t + (1-\alpha)X
\end{equation}
where $1-\alpha$ is the jump probability.
We can obtain the equilibrium representation as 
\begin{equation}\label{EQ:Diffusion}
{Z}^{*} = (1-\alpha)(I - \alpha P)^{-1}X
\end{equation}

Table 1 summarizes the feature diffusion results of the above three models. In additional to the above three models, some other models can also
be explored here ~\cite{donoser2013diffusion,lu2014learning}.
There are three aspects of the above three diffusion models.
(1) They conduct feature diffusion while  preserve the information of original input feature $X$ in feature representation process.
(2) They have explicit optimization formulation. The equilibrium representation of these models  can be obtained via a simple closed-form solution. 
(3) They can  be naturally extended to address  the data with multiple graph structures, as shown in \S 3.

\begin{table}[!htp]
\centering
\caption{Feature diffusion methods}
\centering
\begin{tabular}{c|c}
  \hline
  \hline
  Model & Diffusion function $\mathcal{H}_d(A,X)$  \\
  \hline
  Laplacian diffusion  &\(\alpha(I+\alpha (D-A))^{-1}X\)  \\
  Random walks with restart  &\((1-\alpha)(I-\alpha A D^{-1})^{-1}X\) \\
  Normalized Laplacian diffusion &\((I - \alpha D^{-1/2} A D^{-1/2})^{-1}X\) \\
  \hline
  \hline
\end{tabular}
\end{table}
%

%
%
%
%
%

\subsection{Graph embedding}

Graph embedding techniques have been widely used in dimensionality reduction and label prediction.
Given a graph $G(V, E)$ with adjacency matrix $A$ and $X=(X_1\cdots X_n)\in \mathbb{R}^{n\times d}$. The aim of graph embedding is to
generate a low-dimensional representation $Y_i\in \mathbb{R}^{k}, k< n$ and $Y_i = g_e(X_i)$ for node $v_i$.
One  popular way is to utilize linear embedding, which assumes that 
\begin{equation}\label{EQ:embedding}
Y_i =g_e(X_i) = X_i W \  \ \ \ \ \ \text{or} \ \ \ \ \ Y = g_e(X) = XW
\end{equation}
where $W\in \mathbb{R}^{d\times k} $ denotes the linear projection matrix.

\section{Graph Diffusion-Embedding Networks}

In this section, we present our graph diffusion-embedding networks (GDEN).
Similar to previous GCN ~\cite{defferrard2016convolutional,kipf2016semi}, 
The aim of our GDEN is seek a nonlinear function $f(X,A)$ to conduct dimensionality reduction and label prediction.
It contains several propagation layers and one final perceptron layer together.
For different tasks, one can design different final perceptron layers.

\subsection{Propagation layer} %

Given any input feature $X=X^{(0)}$ and graph structure (adjacency matrix) $A$, GDEN conducts
the  layer-wise propagation rule as, 
\begin{align}
& X^{(k)} = \sigma\big(\mathcal{H}_d(A,X^{(k-1)})W^{(k)}\big)
\end{align}
where $k=1,2\cdots K$. Function $\mathcal{H}_d(A,X^{(k-1)})$ denotes the diffused feature representation.
We can use any kind of diffusion models (Table 1) in our GDEN layer.
Parameter $W^{(k)}\in \mathbb{R}^{d_{k-1}\times d_{k}}$ is a  layer-specific trainable weight matrix which is used to conduct linear projection.
Function $\sigma(\cdot)$ denotes an activation function, such as $\text{ReLU}(\cdot) = \max(0,\cdot)$.
The output $X^{(k)}\in \mathbb{R}^{n\times d_k}$ of the $k$-th layer provides a kind of low-dimensional embedding for graph nodes.

\subsection{Final perceptron layer}

One can design different final perceptron layers for different problems. In the following, we present two kinds of
perceptron layers for semi-supervised learning and link prediction problem, respectively.

\textbf{(1) Semi-supervised learning}


For semi-supervised learning, let ${L}$ indicate the set of labelled nodes and
$Y_{{L}}$ be the corresponding labels for labelled nodes.
The aim of semi-supervised learning is to predict the labels for the unlabelled nodes.
To do so, we can design a final perceptron layer as
\begin{align}
M = \mathrm{softmax} (\mathcal{H}_d(A,X^{(K)}) \widetilde{W} )
\end{align}
where $M\in \mathbb{R}^{n\times c}$ is the label output of the final layer and $M_i$ is the label indication vector of node $v_i$.
Similar to GCN~\cite{kipf2016semi}, we can use the following cross-entropy loss function over all labeled nodes $L$ for semi-supervised classification. 
\begin{equation}
\mathcal{L}_{\text{Semi-GDEN}} = -\sum\nolimits_{i\in L} \sum^c\nolimits_{j=1} Y_{ij}\mathrm{ln} M_{ij}
\end{equation}

\textbf{(2) Graph auto-encoder}

The aim of graph auto-encoder (GAE) is to reconstruct/recover an optimal graph $\tilde{A}$ based on input feature $X$ and initial graph $A$.
Similar to work \cite{kipf2016variational}, we can use GDEN encoder and a simple inner product decoder, which can be used for link prediction task.
For GAE problem, the final perceptron layer can be designed as the inner product of the final output embedding $X^{(k)}$, i.e.,
\begin{equation}
\tilde{A} = \sigma(X^{(K)}{X^{(K)}}^{T})
\end{equation}
where $\sigma$ is the logistic sigmoid function and $X^{(K)}$ is the output representation of the final propagation layer.
One can use MSE loss function here which is  defined as 
 \begin{equation}
\mathcal{L}_{\text{GAE-GDEN}} = \| \tilde{A} - A\|^2_F 
 \end{equation}
Some other loss functions, such as cross-entropy loss, can also be used here.
\begin{figure}[!htb]
\centering
\centering
\includegraphics[width=0.475\textwidth]{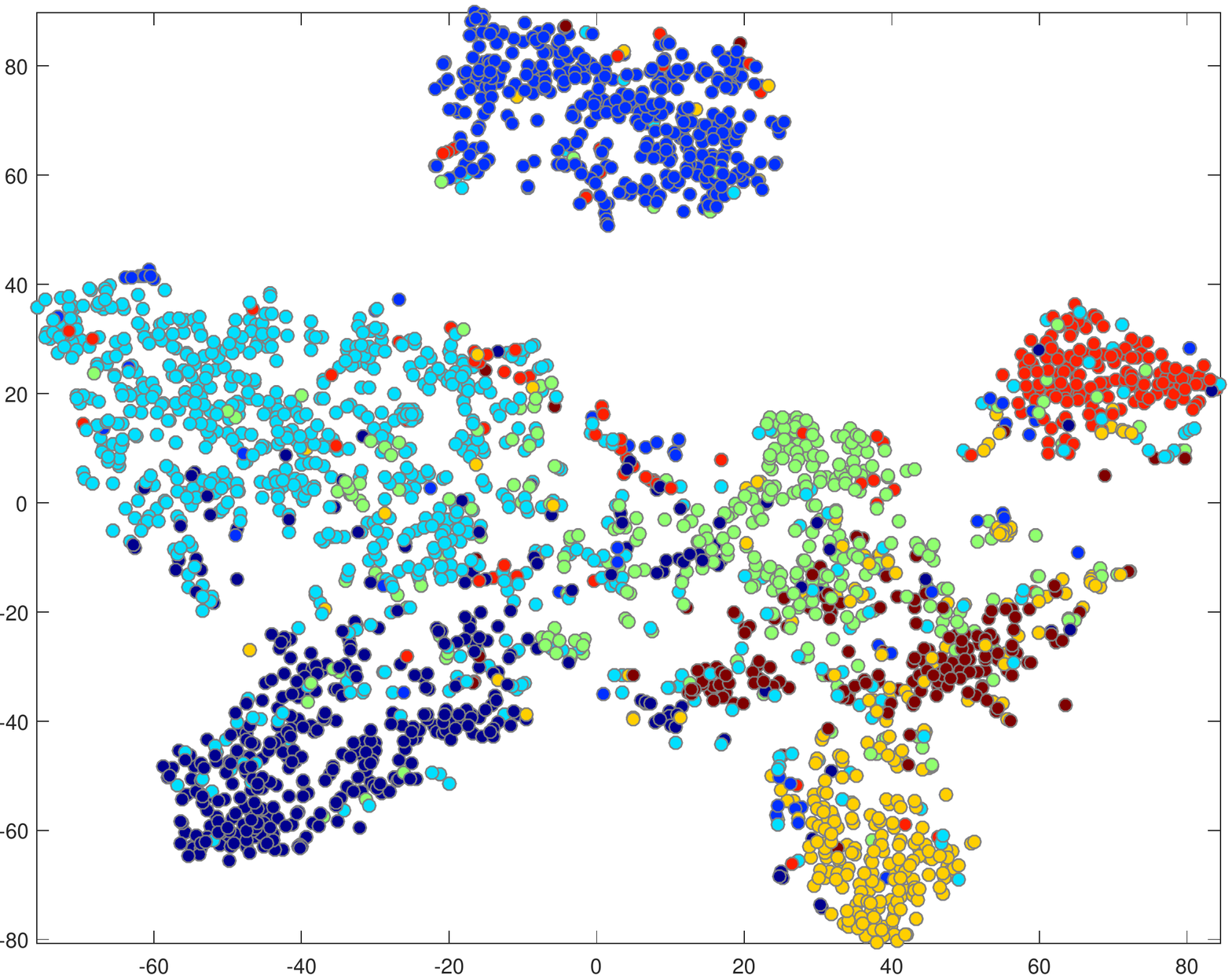} \includegraphics[width=0.475\textwidth]{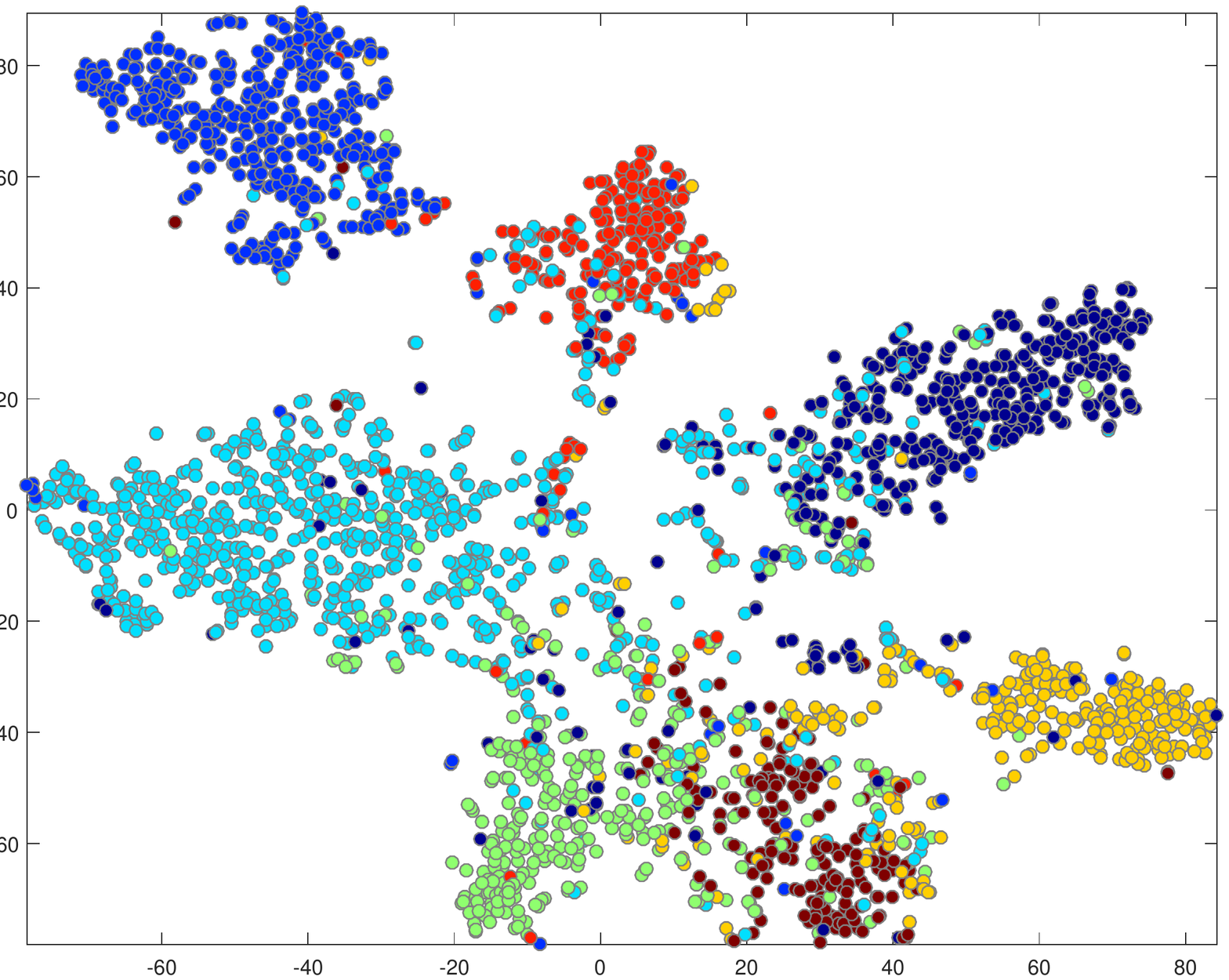}
  \caption{t-SNE plot of the obtained feature representation based on the first hidden layer of GCN~\cite{kipf2016semi} and GDEN respectively on the Cora dataset. Node colors denote classes. (LEFT: GCN result ~\cite{kipf2016semi}
  RIGHT: GDEN result). The data distribution of different classes is demonstrated more clearly in our GDEN representation. }\label{fig::lambda}
\end{figure}
\subsection{Comparison with related works} 

We provide a detail comparisons with recent graph convolutional network (GCN)~\cite{kipf2016semi}, diffusion convolutional recurrent
neural network (DCNN)~\cite{atwood2016diffusion} and recent diffusion convolutional recurrent neural network (DCRNN)~\cite{li2018dcrnn_traffic}.

Previous GCN, DCNN and GCRNN generally use a random walk based diffusion process.
In GCN~\cite{kipf2016semi}, it utilizes a one-step diffusion  while DCRNN~\cite{li2018dcrnn_traffic} utilizes a finite $K$-step truncation of diffusion  on a graph.
One main limitation  is that the equilibrium (convergence) representation of feature diffusion can not be obtained.
Also, these models can not be used directly for the data with multiple graph structures.
In contrast, in our GDEN, we explore \emph{regularized diffusion} models (as shown in Table 1). The benefits are three aspects.
(1) They conduct feature diffusion while  preserve the information of original input feature $X$ in feature representation process.
(2) They have explicit optimization formulation. The equilibrium representation of diffusion in our models can also be obtained via a
simple closed-form solution which can thus be computed efficiently.
(3) They can  be naturally extended to address  the data with multiple graph structures. 

\section{Multi-GDEN}

Comparing with previous (GCN)~\cite{kipf2016semi}, DCNN ~\cite{atwood2016diffusion} and DCRNN ~\cite{li2018dcrnn_traffic}, one benefit of
the proposed GDEN is that it can naturally deal with structured data with multiple graph structures.

Given $X\in \mathbb{R}^{n\times d}$ with multiple graph structures $\{A^{(1)},A^{(2)}\cdots A^{(m)}\}$,
we aim to seek a nonlinear function $f(X; A^{(1)}\cdots A^{(m)})$ to conduct dimensionality reduction and
label prediction. This is known as multiple graph learning problem.

First, for multiple graphs, we can conduct feature diffusion as
\begin{equation}\label{EQ:Diffusion}
\min_{Z} \ \ \frac{1}{2} \sum^m_{v=1}\big[\sum^n_{i,j=1} A^{(v)}_{ij} \|Z_i - Z_j\|^2_F\big] + \alpha \sum^n_{i=1} \|Z_i - X_i\|^2_F
\end{equation}
The optimal closed-form solution for this problem is given by
\begin{equation}\label{EQ:Diffusion}
Z^* = \alpha\big(I + \alpha \sum^m_{v=1} (D^{(v)} - A^{(v)})\big)^{-1}X
\end{equation}
where $D^{(v)} = \text{diag}\{d^{(v)}_{11},d^{(v)}_{22},\cdots d^{(v)}_{nn}\}$ and $d^{(v)}_{ii} = \sum_{j}A^{(v)}_{ij}$, and $D^{(v)} - A^{(v)}$ is
the Laplacian of the $v$-th graph.
Similarly, we can also derive multiple graph feature diffusion based on normalized Laplacian diffusion.

Then, we can thus incorporate this multiple graph feature diffusion in each layer of GDEN (Eqs.(9,10)) to achieve multiple graph diffusion and embedding.

%

\section{Experiments}

To evaluate the effectiveness of the proposed GDEN.
We follow the experimental setup in
work ~\cite{Yang:2016} and test our model on  the citation network datasets including Citeseer, Cora and Pubmed~\cite{sen2008collective}. 
The detail introduction of datasets used in
our experiments are summarized in Table 2.
%
\begin{table}[!htp]
\centering
\caption{Dataset description in experiments}
\centering
\begin{tabular}{c|c|c|c|c|c|c}
  \hline
    \hline
  Dataset & Type & Nodes & Edges & Classes & Features & Label rate \\
  \hline
  Citeseer & Citation network & 3327 & 4732 & 6 & 3703 & 0.036 \\
  Cora & Citation network & 2708 & 5429 & 7 & 1433 & 0.052 \\
  Pubmed & Citation network & 19717 & 44338 & 3 & 500 & 0.003 \\
  \hline
    \hline
\end{tabular}
\end{table}

We compare against the same baseline methods including traditional label propagation
(LP)~\cite{zhu2003semi}, semi-supervised embedding (SemiEmb) ~\cite{weston2012deep}, manifold
regularization (ManiReg)~\cite{belkin2006manifold}, Planetoid ~\cite{Yang:2016}, DeepWalk~\cite{perozzi2014deepwalk} and graph convolutional network (GCN)~\cite{kipf2016semi}.
For GCN~\cite{kipf2016semi}, we implement it using the pythorch code provided by the authors. 
For fair comparison, we also implement our GDEN by using pythorch.
Results for the other baseline methods are taken from work ~\cite{Yang:2016,kipf2016semi}.
We implement it with three versions, i.e.,
1) GDEN-L that utilizes graph Laplacian diffusion in GDEN. 
2) GDEN-RWR that utilizes random walk with restart in GDEN.
3) GDEN-NL that utilizes normalized Laplacian diffusion in GDEN.
The parameter $\alpha$ in GDEN-L, GDEN-RWR and GDEN-NL is set to 4.5, 0.91 and 0.65, respectively.
Table 3 summarizes the comparison results. Here we can note that, 1) GDEN generally performs better than other competing methods, demonstrating
the effectiveness and benefit of the proposed GDEN model.
2) Overall, GDEN-NL performs better than GDEN-L and GDEN-NL.

\begin{table}[!htp]
\centering
\caption{Comparison results of semi-supervised learning on different datasets}
\centering
\begin{tabular}{l|c|c|c}
  \hline
  \hline
  Methond & Citeseer & Cora & Pubmed  \\
  \hline
  ManiReg ~\cite{belkin2006manifold}  & 60.1\% & 59.5\% & 70.7\%  \\
  SemiEmb~\cite{weston2012deep} & 59.6\% & 59.0\% & 71.1\%  \\
  LP~\cite{zhu2003semi} & 45.3\% & 68.0\% & 63.0\%  \\
  DeepWalk~\cite{perozzi2014deepwalk} & 43.2\%  & 67.2\% & 65.3\% \\
  Planetoid~\cite{Yang:2016} & 64.7\% & 75.7\% & 77.2\%  \\
  GCN~\cite{kipf2016semi} & 70.4\% & 81.4\% & 78.6\%  \\
  \hline
  GDEN-L & 71.3\% & 81.9\% & 78.7 \%   \\
  GDEN-RWR & \textbf{72.9\%} &79.3\% &77.9 \%  \\
  GDEN-NL & 72.1\% &\textbf{ 83.0\% }& \textbf{79.2}\%  \\
  \hline
  \hline
\end{tabular}
\end{table}
%

\section{Conclusion}

We  present a novel graph diffusion-embedding networks (GDEN) which operates on graph-structured data.
GDEN integrates both feature diffusion and low-dimensional embedding simultaneously in a unified model.
Based on GDEN, we can easily deal with data with multiple graph structures.
Semi-supervised learning experiments on several  datasets suggest the better performance of the proposed GDEN when
comparing with the recent widely used GCNs.

\bibliographystyle{ieee}
\bibliography{nmfgm}

\end{document}